\documentclass{Interspeech2024}
\usepackage{soul,color,xcolor}

\usepackage{amsmath}
\DeclareMathOperator*{\argmax}{arg\,max}

\usepackage[splitrule]{footmisc}
\interspeechcameraready

\title{Dual-Constrained Dynamical Neural ODEs for Ambiguity-aware Continuous Emotion Prediction}
\name[affiliation={1}]{Jingyao}{Wu}
\name[affiliation={2}]{Ting}{Dang}
\name[affiliation={1}]{Vidhyasaharan}{Sethu}
\name[affiliation={1}]{Eliathamby}{Ambikairajah}

\address{
  $^1$School of Electrical Engineering and Telecommunications, UNSW Sydney, Australia\\
  $^2$University of Melbourne, Australia}
\email{jingyao.wu@unsw.edu.au, ting.dang@unimelb.edu.au, v.sethu@unsw.edu.au, e.ambikairajah@unsw.edu.aui}


\usepackage{cite}

\begin{document}

\maketitle

\begin{abstract}
    There has been a significant focus on modelling emotion ambiguity in recent years, with advancements made in representing emotions as distributions to capture ambiguity. However, there has been comparatively less effort devoted to the consideration of temporal dependencies in emotion distributions which encodes ambiguity in perceived emotions that evolve smoothly over time. Recognizing the benefits of using constrained dynamical neural ordinary differential equations (CD-NODE) to model time series as dynamic processes, we propose an ambiguity-aware dual-constrained Neural ODE approach to model the dynamics of emotion distributions on arousal and valence. In our approach, we utilize ODEs parameterised by neural networks to estimate the distribution parameters, and we integrate additional constraints to restrict the range of the system outputs to ensure the validity of predicted distributions. We evaluated our proposed system on the publicly available RECOLA dataset and observed very promising performance across a range of evaluation metrics.
\end{abstract}

\vspace{-0.3em}
\section{Introduction}
\begin{figure}[t]
  \centering
   \includegraphics[width = 0.4\textwidth, height = 3.3cm]{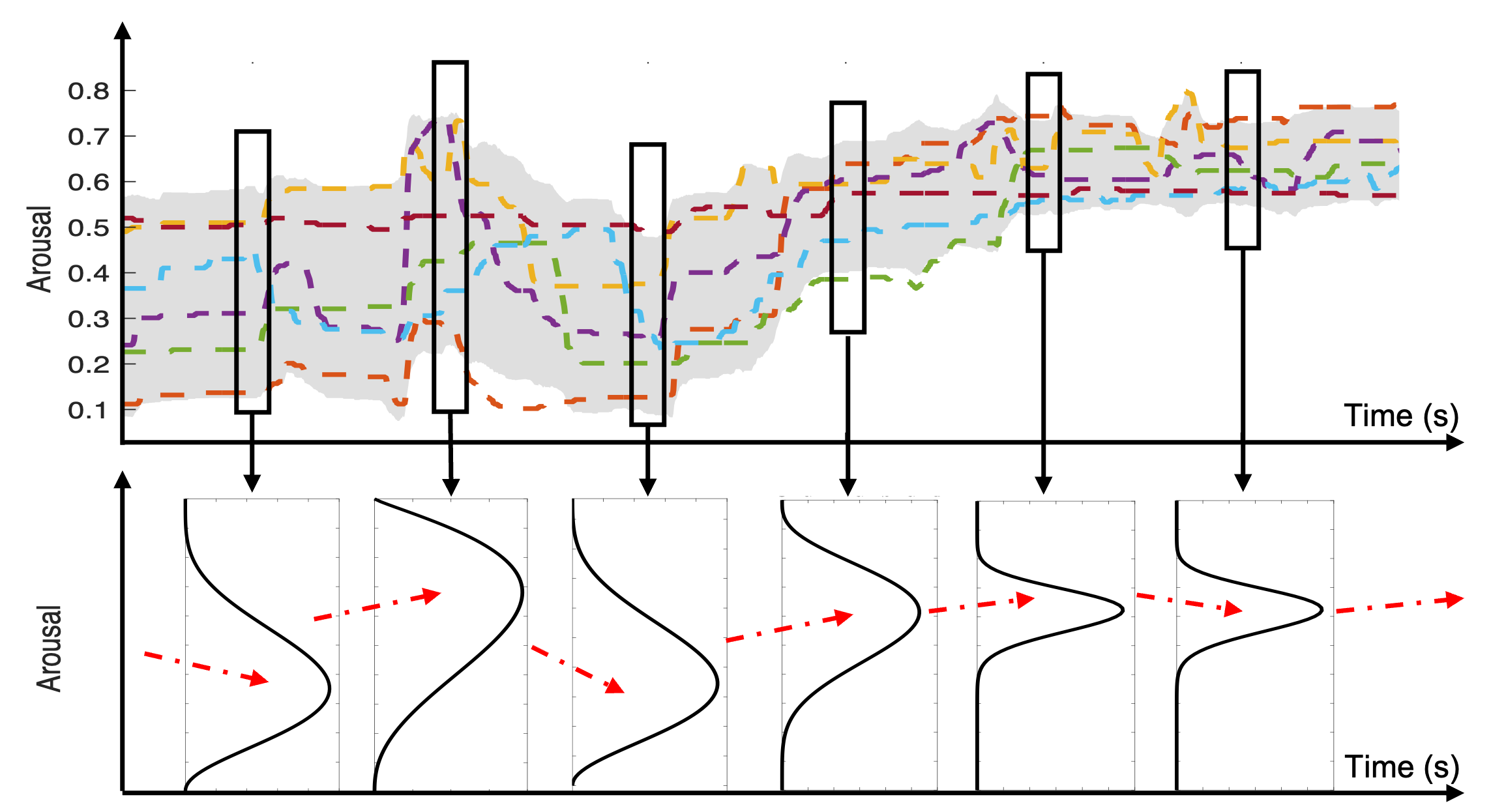}
  \vspace{-1.2em}
  \caption{Illustration of emotion dynamics. The top pane shows six different annotations of the same speech segment over time in coloured lines, and the bottom pane shows a series of distributions reflecting the ambiguous emotion states.}
  \label{fig:fig1}
   \vspace{-2em}
\end{figure}

Speech emotion recognition plays an important role in building the next generation of human-machine interactions \cite{wang2022systematic}. Dimensional representation, describing emotions with continuous values on affect dimensions (e.g., arousal and valence), allows for broader descriptions of the complexity and richness of emotion, making it extensively utilised in the speech emotion recognition research community \cite{russell1980circumplex,gunes2010automatic}. %Thus, this paper also focuses on modelling emotions along the arousal/valence dimensions. %Additionally, emotion is a dynamic process, and a person's emotional state evolves over time \cite{kuppens2017emotion, yadav2022survey}. Continuously tracking the speaker's emotion is appealing, which brings up the more interesting challenge of modelling both time-continuous and value-continuous emotions, leading to Continuous Emotion Recognition (CER) systems \cite{russell1980circumplex,gunes2010automatic}. 
Typically, the continuous emotion labels (also denoted as ratings) are collected from a group of human raters by asking for their perceptual evaluations of the same speech or video recordings. This is depicted in the top plot of Figure \ref{fig:fig1}, where dashed lines represent the various arousal ratings collected from 6 raters. The differences in perception, referred to as \emph{inter-rater ambiguity} \cite{sethu2019ambiguous}, reflect the subtlety of natural emotions, and this ambiguity should be taken into consideration when developing emotion recognition systems.

%Dimensional representation (e.g., arousal and valence) allows broader descriptions of complexity and richness of emotion that represents emotion state with continuous values on affect dimensions, \red{making it extensively employed in the Speech Emotion Recognition (SER) research community. Therefore, this paper will focuses on modelling emotions along the arousal/valence dimensions.} 

%Additionally, emotion is a dynamical process and human's emotional state evloves along time, continuously tracking the speaker's emotion is appealing, which brings up the more interesting challenge of modelling both time-continuous and value-continuous emotions, leading to \emph{Continuous Emotion Recognition} systems \cite{russell1980circumplex,gunes2010automatic}. Typically, emotion ratings (i.e, labels) are collected from a group of human annotators by asking for their perceptual evaluations of the same speech or video recordings, with differences in perception leading to \emph{inter-rater ambiguity} \cite{sethu2019ambiguous}. \red{This is depicted in the top plot of Figure \ref{fig:fig1} that multiple dashed lines indicating the the different arousal ratings annotated by 6 raters.} The perception differences reflects the subtlety of natural emotions and ambiguity should be taken into consideration when developing affective computing systems.

Until recently, researchers in affective computing have often treated this ambiguity as noise and used the average ratings from multiple raters as a `gold standard' \cite{gunes2013categorical, sethu2019ambiguous}. However, there is now growing recognition that inter-rater ambiguity in emotion labels conveying information about the emotion state and such ambiguity is being taken into account by treating arousal/valence labels as distributions ~\cite{dang2017investigation, wu2022novel, mani2021stochastic,zhang2017predicting}. %However, very few works have considered the temporal dependencies of the emotion distributions
This paradigm shift has spurred the development of ambiguity-aware emotion prediction systems that predict emotion states as distributions on the arousal/valence dimensions, rather than conventional mean predictions \cite{dang2017investigation, wu2022novel, mani2021stochastic, zhang2017predicting, han2021exploring, wu2023belief}.
%This in turn has led to the development of a number of ambiguity aware emotion prediction systems that directly model emotion state with distributions on arousal/valence dimensionals, as opposed to the convensional mean predictions ~\cite{dang2017investigation, wu2022novel,mani2021stochastic,zhang2017predicting, han2021exploring}.

Additionally, emotions are dynamic processes that evolve continuously over time \cite{kuppens2017emotion, yadav2022survey}. This dynamic nature applies not only to the emotional state itself but also to the associated ambiguity \cite{dang2018dynamic}. Figure \ref{fig:fig1} illustrates this idea of emotion dynamics with arousal ratings (coloured dashed lines on the top plot) from each rater gradually changing over time. The associated emotion ambiguity (grey shaded area) also follows a smooth evolving process leading to the smoothly varying distributions (bottom plot) that encode the underlying ambiguity. Nevertheless, among the limited works that model emotion ambiguity, only a small subset of these systems have attempted to incorporate the temporal dependencies amongst the distributions.

%There has been an existing effort on developing Continuous Emotion Recognition (CER) systems on affect dimensions, that models both time-continuous and value-continuous emotions.

%Moreover, emotion is a dynamic process, and a person's emotional state evolves over time \cite{kuppens2017emotion, yadav2022survey}. Continuously tracking the speaker's emotion is appealing, which brings up the more interesting challenge of modelling both time-continuous and value-continuous emotions, leading to Continuous Emotion Recognition (CER) systems \cite{russell1980circumplex,gunes2010automatic}. This 

%but only a small subset of these systems also attempt to model the temporal dependencies amongst the emotion labels. Figure \ref{fig:fig1} illustrates this idea of dynamics in emotion with arousal ratings (coloured dashed lines on the top plot) from each rater gradually changing over time. The associated emotion ambiguity (grey shaded area) also follows a smooth evolving process leading to the smoothly varying distributions (bottom plot) that encode the underlying ambiguity. 

In this paper, we present a speech-based ambiguity-aware emotion prediction system that explicitly incorporates the temporal dependencies of emotion distributions. Leveraging the advantages of the recently proposed constrained dynamical neural ordinary differential equation (CD-NODE) for modelling time series with the ability to impose constraints on the smoothness of dynamic process, and its effectiveness in modelling time-varying emotion ratings \cite{dang2023constrained}, we develop a novel ambiguity-aware dual constrained CD-NODE system (denoted as CD-NODE$_\gamma$).
This system predicts ambiguity-aware emotion states with time-varying distributions along the arousal/valence dimensions, and automatically learns how these emotion distributions evolve over time. Moreover, CD-NODE$_\gamma$ integrates smoothness and range constraints to manage the rate of change and the absolute scope of emotion distributions, respectively, further ensuring the validity of the distributions.
%Specifically, the emotion distributions are parameterized as Beta distributions, which have been shown to be well-suited for modelling emotion ambiguity \cite{bose2021parametric}. 

%Beta parameters 
%with both smoothness constraints, controlling the rate of change in emotion distributions, and range constraints, restricting the validity of the distributions.

%that predicts ambiguity aware emotion states with time-varying distributions on arousal/valence dimensions, with the emotion labels parameterised as Beta distributions which have been shown to be well suited for modelling emotion ambiguity \cite{bose2021parametric}. The proposed CD-NODE$_\gamma$ system allows us to directly model the Beta parameters with both smoothness constraint for controlling the rate of temporal variations, and range constraint for  aligning with the requirements of Beta parameterization and matching the original emotion ratings.

%employs a multi-task learning framework that directly models and output .

%CD-NODE based system with dual constraints for ambiguity aware time continuous emotion predictions, with the emotion labels parameterised as Beta distributions which have been shown to be well suited for modelling emotion ambiguity \cite{bose2021parametric}. \jw{claim our novelty more}

\vspace{-0.5em}

%\vspace{-0.3em}
\section{Related work}
%\jw{some literature on modelling distribution, some on mean prediction with temporal dynamics, very limited works taking both,(describ the three) but there is no current llitureture both together}

%\jw{very limited works taking both,(describ the three), other than these three, some literature only modelling distribution, or only on dynamics, but Cd-NODE constraint the smoothess. we don't want the distriution change arbitrily}

Recent years have witnessed an increasing amount of effort on modelling and predicting the ambiguity associated with emotion labels, often treated as probability distributions \cite{dang2017investigation, dang2018dynamic, bose2021parametric}. However, the majority of existing research primarily addresses the ambiguity in predictions while neglecting the temporal dependencies, treating distributions at each time step as independent. Among the few studies addressing temporal dependencies in emotion distributions, initial methods modelled emotion labels using Gaussian Mixture Models (GMMs) without considering temporal dynamics \cite{dang2017investigation}. Subsequent improvements incorporated Kalman filters to account for the temporal evolution of GMM parameters \cite{dang2018dynamic}. However, this approach assumes that the temporal dynamics follow a linear dynamical system. Zhang et al. proposed using Bidirectional LSTM to estimate the means and standard deviations of Gaussian-distributed labels \cite{zhang2018dynamic}, but the poor prediction in standard deviations suggests the unsuitability of the predefined distribution types. While LSTM learns temporal dependencies in the parameters, it lacks transparency and flexibility in integrating constraints or prior knowledge on emotion dynamics. More recent efforts, such as a system utilizing a Sequential Monte Carlo approach, explored predicting time-varying emotion distributions with a non-linear dynamical system, yet focused solely on the dynamics of distribution means \cite{wu2022novel}. These approaches either operate in separate stages rather than end-to-end, apply simplistic linear constraints or make specific and inappropriate assumptions about distribution dynamics, or lack the flexibility in incorporating constraints. On the other hand, the CD-NODE system captures the temporal dynamics of emotions without imposing predefined assumptions on their evolution, facilitating end-to-end modelling and allowing for incorporating constraints, which has shown superior performance in continuous emotion prediction \cite{dang2023constrained}. However, it only predicts mean arousal/valence ratings and neglects the associated ambiguity.

\vspace{-0.5em}
\section{Overview of CD-NODE}

%\vspace{-0.5em}
\begin{figure}[t]
  \centering
  \includegraphics[width = 0.5\textwidth]{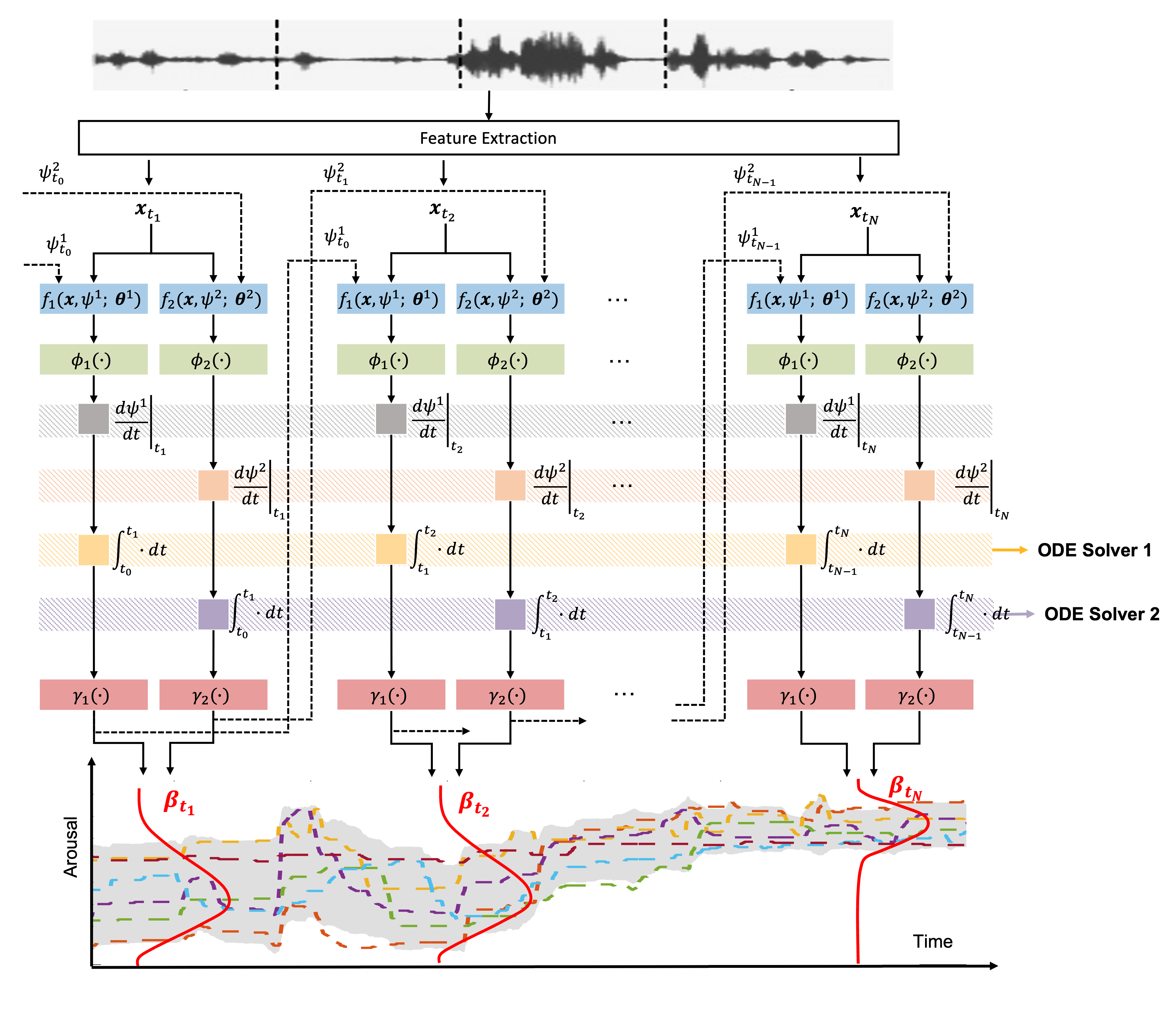}
  \vspace{-3em}
  \caption{Proposed CD-NODE$_\gamma$ for predicting Beta distributions. The speech features $\mathbf{x}_{t_n}$ are fed into two neural networks $f_1$ and $f_2$ which learn the dynamics of each Beta distribution. Rate constraints $\phi_i$ are applied at the outputs of neural networks and range constraints $\gamma_i$ are applied at the outputs of the ODE solvers.}
  \label{fig:2}
  \vspace{-2.3em}
\end{figure}

The recently published CD-NODE was developed to model time series with additional constraints on the dynamics of the predicted quantity. %as an enhancement (variant?) to the original NODE \td{It is not an enhancement or variant. You can directly discuss 'CD-NODE was recently proposed to model time series with additional constraints on dynamics.' }. 
It models the continuous emotion labels (i.e., arousal/valence ratings) $y(t) = \{y_{t_0}, y_{t_1},\cdots, y_{t_N}\}$ as a non-linear dynamical process using ODE with governing function $f$ parameterized by neural networks $\bm \theta$, and the dynamics in $y(t)$ is dependent on both the input speech features $\bm{x}(t) = \{\bm{x}_{t_0}, \bm{x}_{t_1},\cdots, \bm{x}_{t_N}\}$ and previous $y$. It is defined as: %\td{make x all bold to represent vectors? $\theta$ should also be bold.}
\vspace{-0.5em}
\begin{equation} \label{eq:1}
%\small
    \frac{dy(t)}{dt} = \phi(f(\bm {x}(t),y(t); \bm{\theta})))
    \vspace{-0.5em}
\end{equation}
The function $\phi(\cdot)$ is an additional constraint to limit the rate of change of $y(t)$, which controls the  smoothness of $y(t)$ as:
\vspace{-0.8em}
\begin{equation} \label{eq:2}
\vspace{-0.5em}
%\small
    \phi(\cdot) = \alpha * \tanh(\frac{1}{\alpha})
\end{equation}
\noindent where $\alpha$ is a hyperparameter that can be tuned based on the characteristics of $y(t)$. A smaller $\alpha$ results in a smoother trajectory of $y(t)$ and vice versa. The governing function $f$ can be parameterised by any network structure that approximates the nonlinear function in ODE. The final trajectory of $y(t)$ is obtained by solving an ODE initial value problem as shown in Equation \eqref{eq:3}. %In the rest of the paper, we use CD-NODE$_\alpha$ to represent this model.
\vspace{-0.2em}
\begin{equation}\label{eq:3}
%\vspace{-1em}
%\small
      y(t) = \mathrm{ODESolve}(f,\bm\theta,y_{t_0}, t_0, \cdots, t_N, \bm{x}(t)) \\
\end{equation}
%\vspace{-1.8em}

\vspace{-1em}
\section{Proposed CD-NODE$_\gamma$}
%\vspace{-1em}

While the original CD-NODE based system only predicts the mean arousal/valence rating, this paper proposes an ambiguity-aware dual-constrained Neural ODE, referred to as, CD-NODE$_{\gamma}$, that predicts ambiguous arousal/valence states as time-varying Beta distributions as suggested in \cite{bose2021parametric} that Beta distributions are highly suitable for modelling emotion ambiguity. Furthermore, in addition to the explicit smoothness constraint which was a key feature of the CD-NODE, we introduce a second constraint $\gamma$, which restricts the system output to fall within a given range, which in turn helps ensure the validity of the predicted distributions. %\textcolor{red}{Pytorch implementation of CD-NODE$_\gamma$ is made freely available\footnote{Codes will be made publicly available upon publication at Github.}.}

\vspace{-0.5em}
\subsection{Model Structure}
Assuming the arousal/valence at time $t_n$ is represented by a beta distribution that is parameterised by $\bm{\psi}_{t_n}$, the aim of the emotion prediction system is to predict $\bm{\psi}(t) = \{\bm{\psi}_{t_0}, \bm{\psi}_{t_1},\cdots, \bm{\psi}_{t_N}\}$ given a sequennce of input speech features $\bm{x}(t) = \{\bm{x}_{t_0}, \bm{x}_{t_1},\cdots, \bm{x}_{t_N}\}$. The most likely emotion intensity and the associated ambiguity can then be inferred from the predicted beta distribution, $\widehat{\bm{\psi}}_{t_n}$, at each time step.

The Beta distribution is typically parameterised using two parameters, %and such commonly used parameter pairs include shape parameters $\{a,b\}$, \{mean ($\mu$)-variance ($\sigma^2$)\} and \{mode($w$)-concentration($k$)\}. 
denoted as $\bm{\psi}_{t_n} = \{\psi^{1}_{t_n},\psi^{2}_{t_n}\}$. This then leads to a multi-task learning framework to model $\{\psi^{1}_{t_n},\psi^{2}_{t_n}\}$:

\vspace{-0.5em}
\begin{equation}
%\vspace{-0.5em}
\small
    \frac{d\psi^{i}(t)}{dt} = \phi_i(f_i(\mathbf{x}(t),\psi^{i}(t); \bm{\theta}^i)), i \in [1,2]
 %   \vspace{-0.5em}
\end{equation}
\noindent where $i \in [1,2]$ corresponds to the two parameters,  $f_i$ are the two governing functions modelled as neural networks (blue blocks in Figure \ref{fig:2}) with the aim of learning to predict $\frac{d\psi^{i}(t)}{dt}$. They are each constrained with their own smoothness constraint function $\phi_i$ (green blocks in Figure \ref{fig:2}).%\td{rephrased a bit plz check.} 

Noting that the parameters of any beta distributions will have constraints to ensure a resulting distribution is valid (for e.g., when using the mode and concentration parameterisation of a beta distribution, the concentration must be higher than $2$ to ensure a bell shaped distribution), we impose an additional constraint function $\gamma(\cdot)$ to limit the range of $\psi^{1}_{t_n}$ and $\psi^{2}_{t_n}$.
%and more importantly, to regulate the relationship between $\psi^{1}_{t_n}$ and $\psi^{2}_{t_n}$ (red blocks).
Thus, the final outputs are obtained as per equation \eqref{eq:gamma}.
 \begin{equation} \label{eq:gamma}
\small
      \psi^{i}(t) = \gamma_i (\mathrm{ODESolve}(f_i,\bm\theta^i,\psi^{i}_{t_0}, t_0, \cdots, t_N, \bm{x}(t)))%, i \in [1,2] \\
\end{equation}

Finally, the loss function for the proposed multi-task learning framework is taken as: 
\vspace{-0.5em}
\begin{equation} \label{eq:5}
\small
    L = \lambda_1 L_1 + \lambda_2 L_2
\end{equation}
\begin{equation} \label{eq:6}
\small
    L_i= 1 - \rho_c^i(\psi^i(t), \widehat{\psi}^i(t)), i \in [1,2]
\end{equation}

\noindent where $\lambda_i$ are scaling factors; and $L_i$ are loss functions for $\psi^i(t)$, dependent on the concordance correlation coefficient (CCC) $\rho_c^i$ between the predicted $\widehat{\psi}^i(t)$ and the ground truth $\psi^i(t)$~\cite{dang2023constrained}. The model aims to minimize $L_i$, i.e., maximizing the CCC.% $\rho_c^i$ is the concordance correlation coefficient (CCC) between the ground truth $\psi^i(t)$ and predicted Beta parameter $\hat{\psi^i}(t)$:
\vspace{-0.5em}
\subsection{Constraint Function: $\gamma$}

The initial stage of our approach involves identifying the most suitable parameter pairs for the Beta distribution in our proposed CD-NONE$_{\gamma}$. This is a crucial step in ensuring that our model can effectively learn the distributional properties of emotions. The most commonly used parameter pairs include: shape parameters $\{a, b\}$, and $\{mean(\mu), variance(\sigma)\}$ and $\{mode(w),concentration(k)\}$ which indicate the central tendency ($\mu$ or $w$) and spread ($\sigma^2$ or $k$) of the distribution. We observed that shape parameters $\{a, b\}$ are less interpretable compared to $\{\mu, \sigma\}$ and $\{w,k\}$. Furthermore, when comparing $\{\mu, \sigma\}$ to $\{w,k\}$, we observe that both $\mu$ and $\sigma$ are both bounded to the interval $[0,1]$, while only $w$ is bound to a finite interval with the only bound on $k$ being $k > 2$. The lack of an upper bound to $k$ leads to dramatic changes in its dynamics which makes a CD-NODE$_\gamma$ that predicts $\{w,k\}$ harder to train. Therefore, we chose $\{\mu, \sigma\}$ to represent the Beta distributions. Consequently, we design the constraint functions $\gamma_i$ as in equation \eqref{eq:8} for both $\mu$ and $\sigma$ as:

\begin{subequations} \label{eq:8}
\vspace{-2mm}
    \begin{equation} %\label{eq:8}
    \small
    \centering
   % \vspace{-5mm}
     \begin{aligned}
        \gamma_\mu = pS(\cdot), \gamma_\sigma = qS(\cdot) \\
        %\vspace{-3mm}
    \end{aligned}
    \end{equation}
    %\vspace{-7mm}
\begin{equation}\label{eq:8b}
    \small
    \centering
    %\vspace{-1mm}
    \begin{aligned}
       %& ~~~~~~~~~~~~~~~~~~~~~~0 < p \leq 1 \\
       %& 0 < q^2 \leq min(p^2 \frac{1 - p}{1+p}, (1-p)^2 \frac{p}{2-p}) 
    0 < p \leq 1, 0 < q^2 \leq min(p^2 \frac{1 - p}{1+p}, (1-p)^2 \frac{p}{2-p}) 
       %0 < Q^2 \leq P^2 \frac{1 - P}{1+P} \\
       %0 < Q^2 \leq (1-P)^2 \frac{P}{2-P} 
    \end{aligned}
    \end{equation}
%\vspace{-0.5mm}
\end{subequations}
\noindent where, $S(\cdot)$ denotes a sigmoid function and is used to constrain the output to $[0,1]$, and $p$ and $q$ are adjustable constants that must satisfy the relationship as per equation \eqref{eq:8b} in order to ensure the Beta distribution is bell shaped~\cite{bose2024continuous}.

%&&&&&&&&&&&&
\begin{comment}

\vspace{-1em}
\begin{figure}[t]
  \centering
  \includegraphics[width = 0.5\textwidth]{figures/cdnode.png}
  \vspace{-3em}
  \caption{Proposed CD-NODE$_\gamma$ for predicting Beta distributions. The speech features $\mathbf{x}_{t_n}$ are fed into two neural networks $f_1$ and $f_2$ which learn the dynamics of each Beta distribution. Rate constraints $\phi_i$ are applied at the outputs of neural networks and range constraints $\gamma_i$ are applied at the outputs of the ODE solvers.}
  \label{fig:2}
  \vspace{-2.3em}
\end{figure}
%\vspace{-0.5em}
\vspace{0.5em}
\end{comment}
%\vspace{-0.5em}
\section{Experimental Settings}%\td{some titles have comma and some don't. make it consistent.}
%\vspace{-0.5em}
\subsection{Dataset}
The RECOLA dataset is one of the most commonly used multimodal corpus which consists of 9.5 hours spontaneous dyadic conversation recordings in French \cite{ringeval2013introducing}. There are 9 five-minute utterances each in the training and development sets, identical to the data partition in the AVEC challenge 2015 \cite{ringeval2015avec}. Since the challenge test sets are not publicly available, the system is trained and the hyperparameters are optimized using the training set and tested on the development set as per standard practice \cite{dang2023constrained, wu2022novel, bose2024continuous,ringeval2015avec}. The dataset is annotated by 6 human raters with continuous arousal and valence labels within range $[-1,1]$, with sampling rate of 40ms. 

\vspace{-0.5em}
\subsection{Beta distribution parameter estimation}\label{sec:beta}
To obtain the ground truth beta distribution, we follow the same approach used in \cite{bose2024continuous} to estimate the Beta distribution parameters. The original ratings are first mapped to range $x \in [0,1]$ via a linear transformation $y = 0.4975x + 0.5$ in order to perform the Beta fit. To compute the Beta parameter at each frame, the ratings from neighbouring $F$ frames are concatenated as $\widetilde{\bm{y}} = [\bm{y}_{n-F}:\bm{y}_{n+F}]$, $\bm{y}_n$ refers to the labels from 6 raters at frame $n$ to avoid overfitting. $F$ is 6 and 1 for arousal and valence respectively as suggested in \cite{bose2024continuous}. Finally, the Beta parameters $\bm{\psi}_{t_n}$ are obtained by Maximum A Posterior (MAP) as:
\vspace{-0.5em}
\begin{equation}
\small
%\label{eq:beta} \label{eq:infer}
    \bm{\psi}_{t_n} = \argmax_{\bm{\psi}} \prod_{m} P(y_{n,m}|\bm{\psi})P(\bm{\psi})
    \vspace{-2.5mm}
\end{equation}
%\td{$p(y_{n,m}|\psi_n)P(\psi_n)$? also $\argmax_{\psi}_n$} \jw{I think the current equation is correct}\td{not sure. if there is no subscript, it shouldn't be for any of it, right? if there is subscripts, it should be $t_n$ but not $n$? }\jw{I think the $\psi$ in $P(\psi)$ and $P(y_{n,m}| \psi)$ is a random variable, so no subscript}
\noindent where $y_{n,m}$ denotes the rating from the $m^{th}$ rater at frame $n$; $p(y_{n,m}|\psi)$ refers to the posterior probability of $y_{n,m}$; and $P(\psi)$ refers to the prior probability which is estimated with kernel density estimation, the same as in \cite{bose2024continuous}.

\vspace{-0.5em}
\subsection{System configurations}
\vspace{-0.3em}
\noindent\textit{\textbf{Features.}} The Bag-of-audio-words(BoAW) feature representations are extracted with 100 clusters from 20-dimensional MFCCs using OpenXbow \cite{schmitt2017openxbow}. Detailed explanation refers to \cite{schmitt2016border}. 4 second delay compensation is applied to both arousal and valence to compensate the annotation delay \cite{huang2015investigation}. 

\noindent\textit{\textbf{Network parameters. }}%structure}} 
A fully connected (FC) layer with 64 neurons and $tanh$ activation function is first employed to transfer the BoAW features. The following $f_1$ (i.e., $f_\mu$) and $f_2$(i.e., $f_\sigma$) for modelling dynamics of $\mu$ and $\sigma$ are composed of three FC layers respectively, with $tanh$ activation function applied for the first two layers to enforce the nonlinearity, and the smoothing constraint applied to the last FC layer (c.f. Equation \eqref{eq:8}), with $\alpha_u=0.5$~\cite{dang2023constrained} and $\alpha_\sigma=10$ which is optimized within $\{0.5, 1, 2, 4, 6, 8, 10, 15\}$ for both arousal and valence. For the proposed constraints in Equations~\eqref{eq:2}, we chose $p = 0.75$ for both arousal and valence %, which is smaller than 1 and 
as 95\% of the ratings are within the range [0,0.75], and $q = 0.15$ satisfying the requirement in Equation \eqref{eq:8}. The initial values $\psi^{i}_{t_0}$ in Equation \eqref{eq:gamma} are the Beta distribution parameters inferred from the labels of the first frame during the training phase, and set to zeros during the test phase given emotions at the beginning of any utterances tend to be zeros.
%\textcolor{red}{The initial values $\psi^{i}_{t_0}$ in Equation \eqref{eq:gamma} in the training phase are the Beta parameters computed using the labels from the first frame after delay compensation; %following Equation \eqref{eq:infer}; 
%these for test phase are set to be zero since emotions at the beginning of any utterances tend to be zeros.} 
We used the adjoint method with Runge-Kutta 5 ODE solver~\cite{chen2018neural}. The absolute and relative error tolerances of the ODE solver were %chosen as 
$10^{-13}$ and $10^{-7}$ based on preliminary empirical analyses~\cite{dang2023constrained}. The number of model parameters is approximately 30k\footnote{Code: https://github.com/JingyaoWU66/CD-NODE-gamma}.

\noindent\textit{\textbf{Baseline. }}% LSTM system structure
 Two state-of-art systems are compared: the Sequential Monte Carlo (SMC) approach \cite{wu2022novel} and the LSTM based system \cite{bose2024continuous}. The LSTM system follows the same settings as in \cite{bose2024continuous} which directly predicts Beta distribution parameters. In our experiments, the sigmoid activation function is applied at the output layer to limit the output range for a fair comparison. %The LSTM baseline consists of 2 layers with 64 neurons for feature transformation and is followed by two separate FC layers to learn beta distribution parameters, respectively. The sigmoid activation function is applied at the output layer to limit the output range for a fair comparison. 
The SMC approach predicts distributions that are not constrained to Beta distributions and we employ the same settings as in \cite{wu2022novel}.

\noindent\textit{\textbf{Training. }}For all systems, Adam optimizer was used, and the initial learning rate is optimized to 0.01 and 0.001 for $f_\mu$ and $f_\sigma$, with a decaying ratio of 0.9. 60 and 30 epochs were tested for CD-NODE and LSTM baseline. We fixed the seed numbers to ensure reproducibility when testing different models.
%\textcolor{red}{The models are tested using 10 random seeds and the best results are reported with p-value $<$ 0.001. We also fixed the seed numbers to ensure reproducibility when testing different models.} \td{if 10 seedsm reporting best is not the right way. You should report the mean. }
The training utterances were chunked into 4-second windows (i.e., 100 frames) in each batch for training efficiency, while it was tested with the entire utterance to match the practical scenarios. The scaling factors in the loss function in Equation (\ref{eq:5}) were optimized within the range [0,20] with step size 1 and $\lambda_1 = 1$ and $\lambda_2 = 10$ were selected. The predictions were post-processed with a moving average filter with 12 frames(~0.5s) window size. The training time is approximately 72h on Apple M2 chip with 8 cores on 16GB CPU.

\noindent\textit{\textbf{Evaluation Metric. }}It is suggested in \cite{wu2022novel} that at low ambiguity regions, a well-predicted distribution should be centred at the ground truth mean; whereas at high ambiguity regions, the mean is not representative of the underlying emotions, and thus, the spread of the distributions is more important. We evaluate the first aspect by comparing root mean squared error (RMSE) between the predicted mean and ground truth mean at different ambiguity regions \cite{wu2022novel}; and the second aspect using Concordance Correlation Coefficient (CCC) between the predicted standard deviation (SD) and ground truth SD. %\textcolor{red}{We also report the CCC comparison between predicted mean and ground truth mean as per literature \cite{wu2022novel, bose2024continuous}, however, it should be noted that this completely ignores
%ambiguity.

%\vspace{-0.5em}
\begin{table}[t!]
\vspace{-1em}
\vspace{-0.3em}
\scriptsize
\caption{CCC comparison of the predicted Beta distributions in terms of mean and standard deviations (SD).}
%\vspace{-0.5em}%CD-NODE$_{\gamma}$ refers ambiguity-aware CD-NODE \jw{name} configured with both rate and range constraints.
\vspace{-0.7em}
\centering
\begin{tabular}{cllll}
\hline
\textbf{Systems}                & \multicolumn{2}{c}{\textbf{Arousal}}                                                    & \multicolumn{2}{c}{\textbf{Valence}}                                \\ \hline
                                & \multicolumn{1}{c}{\textbf{Mean}}          & \multicolumn{1}{c}{\textbf{SD}}            & \multicolumn{1}{c}{\textbf{Mean}} & \multicolumn{1}{c}{\textbf{SD}} \\
SMC \cite{wu2022novel}                             & {0.702} & {0.403} & 0.391                    & \textbf{0.195}                  \\
\multicolumn{1}{l}{LSTM$-w-k$ \cite{bose2024continuous}} & {0.637} & {0.436} & 0.321                             & 0.044                           \\
LSTM$-\mu-\sigma$                           & 0.658                             & \textbf{0.468}                             & 0.346                             & 0.013                           \\
Proposed CD-NODE$_{\gamma}$                        & \textbf{0.747}                             & 0.412                            & \textbf{0.425}                    & 0.081                  \\ \hline
\end{tabular}
\label{table:1}
%\vspace{-3em}
\end{table}

%, defined as:
%\vspace{-3mm}
\begin{comment}
\begin{align}
% \begin{equation}
 \small
         \rho_c = \frac{2 \rho \sigma_y \sigma_{\hat{y}}}{\sigma_y^2 + \sigma_{\hat{y}}^2 + (\mu_y - \mu_\hat{y})^2} 
     \vspace{-1em}
      \vspace{-2mm}
%\end{equation}
\end{align}

\noindent where $\sigma _{y}$ and $\sigma _{\hat{y}}$ refers to the standard deviation for the ground truth and the predicted quantities, $\mu _{y}$ and $\mu _{\hat{y}}$ are the corresponding mean of the two variables. $\rho$ is the Pearson's correlation coefficient between the two variables.
\vspace{-1em}
\end{comment}

\vspace{-1em}
\section{Results}
%\vspace{-0.5em}
\subsection{Comparisons with state-of-the-art}
%\vspace{-1mm}

The comparison to the state-of-the-art systems is reported in Table~\ref{table:1}, in terms of CCC between the predicted Beta distribution (i.e., mean $\mu$ and standard deviation $\sigma$) and the ground truth Beta distributions (refer to section \ref{sec:beta}) correspondingly. In addition to \cite{bose2024continuous} that uses LSTM to predict $w$ and $k$ of beta distribution (referred to as LSTM$-w-k$), we further validate the system performance for predicting $\mu$ and $\sigma$ for a fair comparison, named as LSTM$-\mu-\sigma$. The proposed system is also compared to the existing SMC model which adopts non-parametric distribution~\cite{wu2022novel}. It is observed that the proposed model CD-NODE$_{\gamma}$ %\jw{name} \td{looks like lstm systems are named with the output variable such as $u or w k$, but CD-NODE is named with the constraint? it is better to make them consistent.} \jw{Yeah, but is it okay to name it like this? I mean maybe too many symbols?}\td{yes agreed. Maybe don't use $\alpha$ but only $beta$, and also use only $u$ instead of $u$ and $\sigma$.} 
yields the highest CCC in terms of mean predictions $\mu$ for both arousal and valence, with significance p-value $<$ 0.001. Regarding the SD prediction, CD-NODE$_{\gamma}$ shows comparable performance to the best system LSTM$-\mu-\sigma$ for arousal, and underperforms the best system SMC for valence. This could be because the dynamics of SD change more rapidly compared to the mean, which may require a more advanced ODE solver to obtain a more accurate solution.

%\vspace{-0.5em}
\begin{figure}[t!]
\vspace{-1.8em}
    \centering
    \small
    \includegraphics[width = 0.4\textwidth, height = 4.7cm]{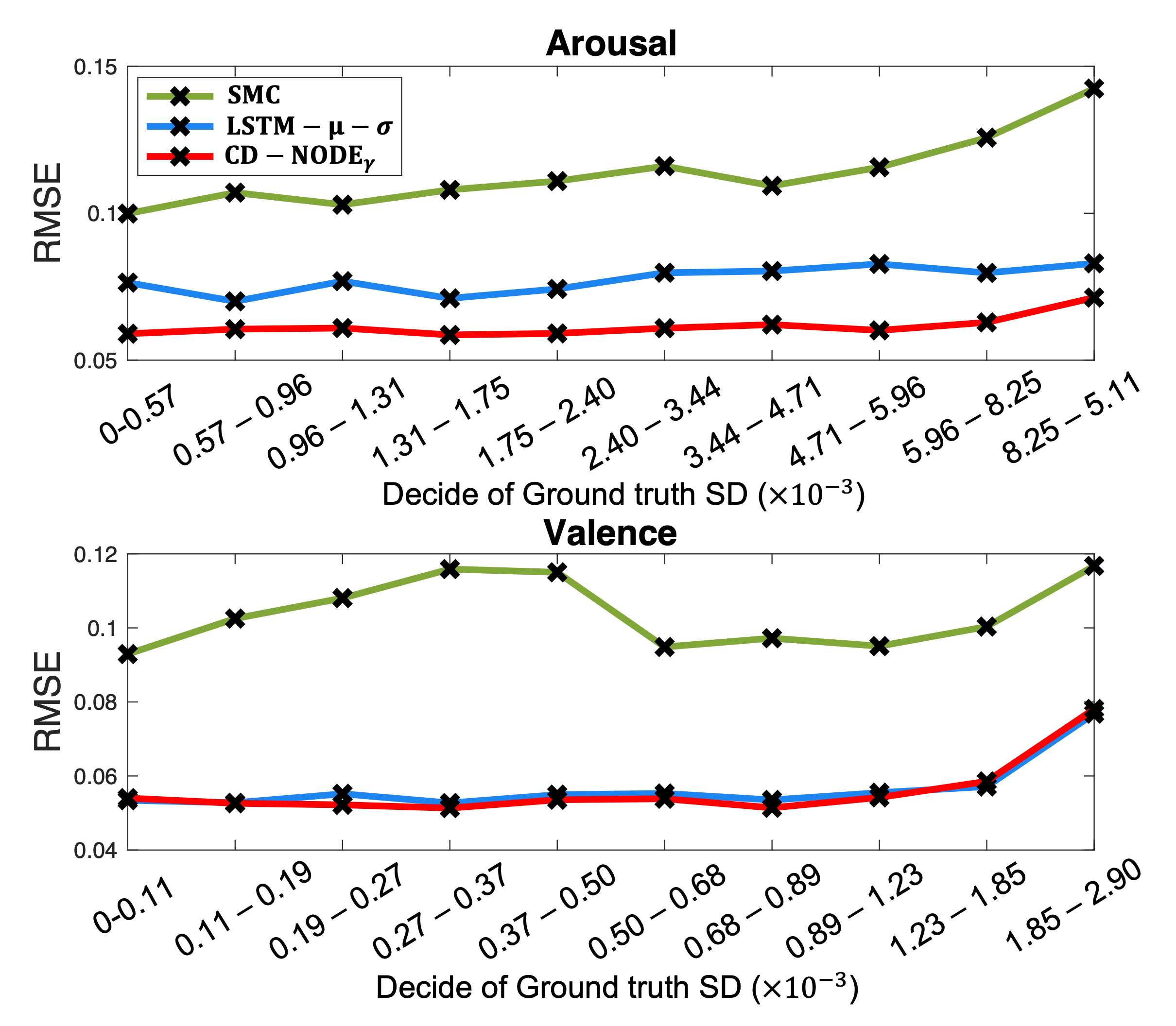}
    \vspace{-1em}
    \caption{RMSE of the proposed CD-NODE$_\gamma$ and baselines. The standard deviation (SD) range corresponding to each decile is shown on the x-axis.} 
    \label{fig:mse1}
\vspace{-2.5em}
\end{figure}

Figure \ref{fig:mse1} presents the RMSE values between the predicted and ground truth means across different ambiguity regions, defined based on deciles of the ground truth standard deviation. As expected, the RMSE generally increases at higher standard deviation deciles for both arousal and valence, indicating that mean prediction errors increase with rising ambiguity (low inter-rater agreement). Conversely, the systems perform better at low ambiguity regions (high inter-rater agreement), which supports our hypothesis. Notably, our proposed system, CD-NODE${_\gamma}$, consistently achieves the lowest RMSE compared to all baseline systems for both arousal and valence. %\textcolor{red}{Although CD-NODE$_{\gamma}$ and LSTM$-\mu-\sigma$ achieved similar performance, this may be attributed to the inherent difficulty in predicting valence using speech data. However, it's worth noting that CD-NODE$_{\gamma}$ still significantly outperformed SMC on valence.} %\textcolor{red}{For instance, the RMSE performance of CD-NODE${\gamma}$ at the lowest ambiguity decile achieved relative decrease by 40\% and 52.6\% than that of SMC on arousal and valence respectively. Similarly, that of CD-NODE${\gamma}$ relatively dropped by 25\% compared to LSTM$-\mu-\sigma$ on arousal. While it achieved similar performance on valence, which might be because valence is generally worse predicted using speech.} 
These results demonstrates the superior performance of CD-NODE${\gamma}$ in modelling time-varying emotion distributions, as it accurately predicts the emotion state at low ambiguity regions while still reasonably predicting the level of ambiguity at both low and high ambiguity regions. Although CD-NODE$_{\gamma}$ and LSTM$-\mu-\sigma$ achieved similar performance on valence, it may be attributed to the inherent difficulty in predicting valence using speech data.

\vspace{-0.5em}
\subsection{Impact of range constraint}
\vspace{-0.2em}

We further validate the effectiveness of the constraint functions in the CD-NODE$_\gamma$, by comparing the system performance without constraint (D-NODE), with only the smoothness constraint (CD-NODE), and with both the smoothness and range constraints (CD-NODE$_\gamma$). The performance in terms of CCC in Table \ref{table:ccc2} shows that CD-NODE$_\gamma$ achieves the highest CCC and D-NODE without constraint shows the lowest CCC, indicating the advantages of adding constraints. More importantly, CD-NODE$_\gamma$ yields 3.6\% and 6.5\% relative improvements over CD-NODE for mean and SD of arousal,  and 3.7\% and 26.6\% relative improvements for mean and SD of valence. This further confirms the effectiveness of the proposed range constraint $\gamma$.

\begin{figure}[tb!]
\vspace{-0.8em}
    \centering
    \small
    \includegraphics[width = 0.4\textwidth, height = 4.7cm]{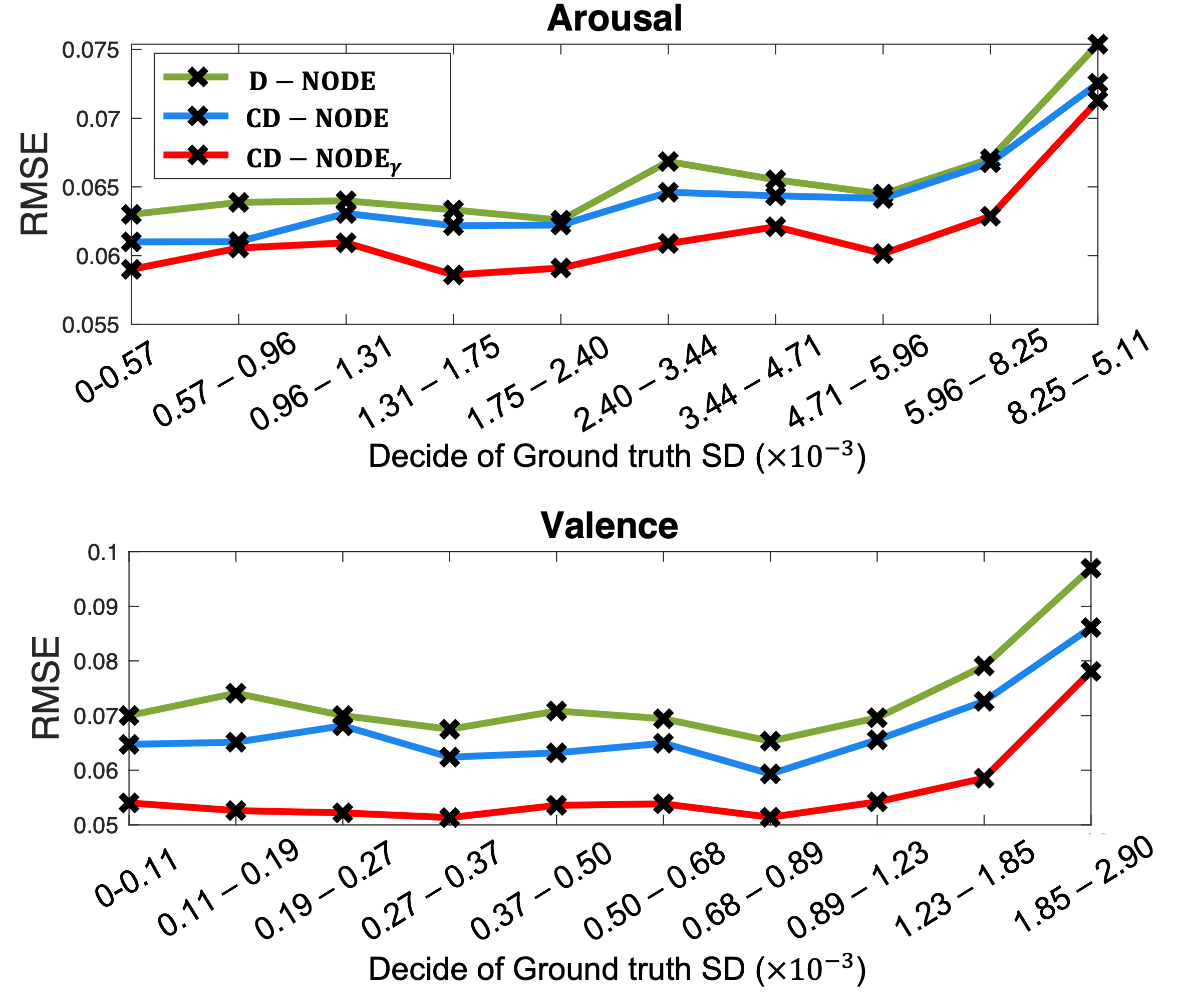}
    \vspace{-1.3em}
    \caption{RMSE of the proposed CD-NODE$_\gamma$, and baselines.} %The standard deviation (SD) range corresponding to each decile is shown on the x-axis.} 
    \label{fig:mse2}
    \vspace{-2em}
\end{figure}

RMSE is also reported in Figure \ref{fig:mse2} for the three systems. All three systems show an increasing trend of RMSE with increasing SD decibels, aligning with the assumption that prediction error is larger where inter-rater ambiguity is high. Notably, CD-NODE$_\gamma$ consistently shows a smaller RMSE compared to the other two systems, e.g., 6.3\% and 3.3\% relative decreases at lowest ambiguity decile on arousal compared to D-NODE and CD-NODE respectively, further validating the benefits of proposed constraints. Overall, the results suggest that CD-NODE$_\gamma$ with the range constraint can capture the time-varying dynamics of emotion distributions more effectively, which might be because the constraint function eliminates the noisy predictions outside the possible range of the predicted quantity.

%\begin{comment}
    
%\end{comment}
\vspace{-0.5em}
\begin{table}[tb!]
\vspace{-1em}
\vspace{-0.3em}
\scriptsize
\centering
\caption{CCC comparison of the three systems: D-NODE without constraint, CD-NODE with rate constraint and CD-NODE$_{\gamma}$ with rate and range constraint.}
\vspace{-0.5em}
%\begin{adjustbox}{0.8\columnwidth,center}
%\resizebox{\columnwidth}{!}{\begin{tabular}{ccccc}
\begin{tabular}{ccccc}
\hline
\textbf{Systems}  & \multicolumn{2}{c}{\textbf{Arousal}} & \multicolumn{2}{c}{\textbf{Valence}} \\ \hline
    & \textbf{Mean} & \textbf{SD} & \textbf{Mean} & \textbf{SD}     \\
D-NODE &     0.703 &     0.359&  0.397& 0.058\\
CD-NODE & 0.721  & 0.387 &  0.410 &    0.064 \\
{Proposed CD-NODE$_{\gamma}$} & {\textbf{0.747}} & {\textbf{0.412}} & \textbf{0.425}      &   \textbf{0.081}         \\ \hline
\end{tabular}%}
%\vspace{-3em}
%\end{adjustbox}
\label{table:ccc2}
\end{table}

%\vspace{-0.2em}
\section{Conclusion}
%\vspace{-0.3em}
In this paper, a multi-task learning framework is used to develop an ambiguity-aware dual-constrained dynamic neural ODE (CD-NODE$_\gamma$) system that models ambiguous emotional states as time-varying Beta distributions. Our proposed method takes into account the temporal dynamics of distributions and incorporates a range constraint that restricts the outputs to better ensure the validity of predicted emotion distribution, in addition to a smoothness constraint. Experimental results, when compared to various baselines, show that the proposed CD-NODE$_\gamma$ model outperforms them in modelling emotional states and produces accurate predictions in low ambiguity regions and reliable predictions of high ambiguity regions. Moreover, the outcomes acquired with various constraint configurations indicate that the inclusion of a range constraint enhances the CD-NODE's ability to learn the distribution dynamics. The results highlight the benefits of utilising CD-NODE$_\gamma$ to model temporal dynamics of distributions with explicit constraints, which opens up new avenues for emotion ambiguity modelling.

\vspace{-0.5em}

\bibliographystyle{IEEEtran}
\bibliography{mybib}

\begin{thebibliography}{10}
\providecommand{\url}[1]{#1}
\csname url@samestyle\endcsname
\providecommand{\newblock}{\relax}
\providecommand{\bibinfo}[2]{#2}
\providecommand{\BIBentrySTDinterwordspacing}{\spaceskip=0pt\relax}
\providecommand{\BIBentryALTinterwordstretchfactor}{4}
\providecommand{\BIBentryALTinterwordspacing}{\spaceskip=\fontdimen2\font plus
\BIBentryALTinterwordstretchfactor\fontdimen3\font minus \fontdimen4\font\relax}
\providecommand{\BIBforeignlanguage}[2]{{%
\expandafter\ifx\csname l@#1\endcsname\relax
\typeout{** WARNING: IEEEtran.bst: No hyphenation pattern has been}%
\typeout{** loaded for the language `#1'. Using the pattern for}%
\typeout{** the default language instead.}%
\else
\language=\csname l@#1\endcsname
\fi
#2}}
\providecommand{\BIBdecl}{\relax}
\BIBdecl

\bibitem{wang2022systematic}
Y.~Wang, W.~Song, W.~Tao, A.~Liotta, D.~Yang, X.~Li, S.~Gao, Y.~Sun, W.~Ge, W.~Zhang \emph{et~al.}, ``A systematic review on affective computing: Emotion models, databases, and recent advances,'' \emph{Information Fusion}, 2022.

\bibitem{russell1980circumplex}
J.~A. Russell, ``A circumplex model of affect.'' \emph{Journal of personality and social psychology}, vol.~39, no.~6, p. 1161, 1980.

\bibitem{gunes2010automatic}
H.~Gunes and M.~Pantic, ``Automatic, dimensional and continuous emotion recognition,'' \emph{International Journal of Synthetic Emotions (IJSE)}, vol.~1, no.~1, pp. 68--99, 2010.

\bibitem{sethu2019ambiguous}
V.~Sethu, E.~M. Provost, J.~Epps, C.~Busso, N.~Cummins, and S.~Narayanan, ``The ambiguous world of emotion representation,'' \emph{arXiv preprint arXiv:1909.00360}, 2019.

\bibitem{gunes2013categorical}
H.~Gunes and B.~Schuller, ``Categorical and dimensional affect analysis in continuous input: Current trends and future directions,'' \emph{Image and Vision Computing}, vol.~31, no.~2, pp. 120--136, 2013.

\bibitem{dang2017investigation}
T.~Dang, V.~Sethu, J.~Epps, and E.~Ambikairajah, ``An investigation of emotion prediction uncertainty using gaussian mixture regression.'' in \emph{INTERSPEECH}, 2017, pp. 1248--1252.

\bibitem{wu2022novel}
J.~Wu, T.~Dang, V.~Sethu, and E.~Ambikairajah, ``A novel sequential monte carlo framework for predicting ambiguous emotion states,'' in \emph{ICASSP 2022-2022 IEEE International Conference on Acoustics, Speech and Signal Processing (ICASSP)}.\hskip 1em plus 0.5em minus 0.4em\relax IEEE, 2022, pp. 8567--8571.

\bibitem{mani2021stochastic}
T.~Mani~Kumar, E.~Sanchez, G.~Tzimiropoulos, T.~Giesbrecht, and M.~Valstar, ``Stochastic process regression for cross-cultural speech emotion recognition,'' \emph{Proc. Interspeech 2021}, pp. 3390--3394, 2021.

\bibitem{zhang2017predicting}
B.~Zhang, G.~Essl, and E.~Mower~Provost, ``Predicting the distribution of emotion perception: capturing inter-rater variability,'' in \emph{Proceedings of the 19th ACM International Conference on Multimodal Interaction}, 2017, pp. 51--59.

\bibitem{han2021exploring}
J.~Han, Z.~Zhang, Z.~Ren, and B.~Schuller, ``Exploring perception uncertainty for emotion recognition in dyadic conversation and music listening,'' \emph{Cognitive Computation}, vol.~13, pp. 231--240, 2021.

\bibitem{wu2023belief}
J.~Wu, T.~Dang, V.~Sethu, and E.~Ambikairajah, ``Belief mismatch coefficient (bmc): A novel interpretable measure of prediction accuracy for ambiguous emotion states,'' in \emph{2023 11th International Conference on Affective Computing and Intelligent Interaction (ACII)}.\hskip 1em plus 0.5em minus 0.4em\relax IEEE, 2023, pp. 1--8.

\bibitem{kuppens2017emotion}
P.~Kuppens and P.~Verduyn, ``Emotion dynamics,'' \emph{Current Opinion in Psychology}, vol.~17, pp. 22--26, 2017.

\bibitem{yadav2022survey}
S.~P. Yadav, S.~Zaidi, A.~Mishra, and V.~Yadav, ``Survey on machine learning in speech emotion recognition and vision systems using a recurrent neural network (rnn),'' \emph{Archives of Computational Methods in Engineering}, vol.~29, no.~3, pp. 1753--1770, 2022.

\bibitem{dang2018dynamic}
T.~Dang, V.~Sethu, and E.~Ambikairajah, ``Dynamic multi-rater gaussian mixture regression incorporating temporal dependencies of emotion uncertainty using kalman filters,'' in \emph{2018 IEEE International Conference on Acoustics, Speech and Signal Processing (ICASSP)}.\hskip 1em plus 0.5em minus 0.4em\relax IEEE, 2018, pp. 4929--4933.

\bibitem{dang2023constrained}
T.~Dang, A.~Dimitriadis, J.~Wu, V.~Sethu, and E.~Ambikairajah, ``Constrained dynamical neural ode for time series modelling: A case study on continuous emotion prediction,'' in \emph{ICASSP 2023-2023 IEEE International Conference on Acoustics, Speech and Signal Processing (ICASSP)}.\hskip 1em plus 0.5em minus 0.4em\relax IEEE, 2023, pp. 1--5.

\bibitem{bose2021parametric}
D.~Bose, V.~Sethu, and E.~Ambikairajah, ``Parametric distributions to model numerical emotion labels,'' \emph{Proc. Interspeech 2021}, pp. 4498--4502, 2021.

\bibitem{zhang2018dynamic}
Z.~Zhang, J.~Han, E.~Coutinho, and B.~Schuller, ``Dynamic difficulty awareness training for continuous emotion prediction,'' \emph{IEEE Transactions on Multimedia}, vol.~21, no.~5, pp. 1289--1301, 2018.

\bibitem{bose2024continuous}
D.~Bose, V.~Sethu, and E.~Ambikairajah, ``Continuous emotion ambiguity prediction: Modeling with beta distributions,'' \emph{IEEE Transactions on Affective Computing}, no.~01, pp. 1--12, 2024.

\bibitem{ringeval2013introducing}
F.~Ringeval, A.~Sonderegger, J.~Sauer, and D.~Lalanne, ``Introducing the recola multimodal corpus of remote collaborative and affective interactions,'' in \emph{2013 10th IEEE international conference and workshops on automatic face and gesture recognition (FG)}.\hskip 1em plus 0.5em minus 0.4em\relax IEEE, 2013, pp. 1--8.

\bibitem{ringeval2015avec}
F.~Ringeval, B.~Schuller, M.~Valstar, R.~Cowie, and M.~Pantic, ``Avec 2015: The 5th international audio/visual emotion challenge and workshop,'' in \emph{Proceedings of the 23rd ACM international conference on Multimedia}, 2015, pp. 1335--1336.

\bibitem{schmitt2017openxbow}
M.~Schmitt and B.~Schuller, ``Openxbow: introducing the passau open-source crossmodal bag-of-words toolkit,'' 2017.

\bibitem{schmitt2016border}
M.~Schmitt, F.~Ringeval, and B.~W. Schuller, ``At the border of acoustics and linguistics: Bag-of-audio-words for the recognition of emotions in speech.'' in \emph{Interspeech}, 2016, pp. 495--499.

\bibitem{huang2015investigation}
Z.~Huang, T.~Dang, N.~Cummins, B.~Stasak, P.~Le, V.~Sethu, and J.~Epps, ``An investigation of annotation delay compensation and output-associative fusion for multimodal continuous emotion prediction,'' in \emph{Proceedings of the 5th International Workshop on Audio/Visual Emotion Challenge}, 2015, pp. 41--48.

\bibitem{chen2018neural}
R.~T. Chen, Y.~Rubanova, J.~Bettencourt, and D.~K. Duvenaud, ``Neural ordinary differential equations,'' \emph{Advances in neural information processing systems}, vol.~31, 2018.

\end{thebibliography}
\end{document}